\documentclass[11pt,a4paper]{article}
\usepackage[hyperref]{ranlp2021}
\usepackage{times}
\usepackage{latexsym}
\usepackage[utf8]{inputenc}
\usepackage{textcomp,csquotes,hyperref,microtype,times,tabularx,latexsym,qtree,graphicx,verbatim,xcolor,listings,multirow}
\usepackage[english]{babel}
\usepackage{microtype}

\aclfinalcopy 

\title{
SocialVisTUM: An Interactive Visualization Toolkit for \\ 
Correlated Neural Topic Models on Social Media Opinion Mining
}

\author{
Gerhard Hagerer\textsuperscript{{\normalfont 2}} \and 
Martin Kirchhoff\textsuperscript{{\normalfont ~3}} \and 
Hannah Danner\textsuperscript{{\normalfont 2}} \and
Robert Pesch\textsuperscript{{\normalfont 3}}  \\
{\bf Mainak Ghosh}\textsuperscript{1} \and 
{\bf Archishman Roy}\textsuperscript{2} \and 
{\bf Jiaxi Zhao}\textsuperscript{2} \and 
{\bf Georg Groh}\textsuperscript{2} \\
\textsuperscript{1}Max Planck Institute for Innovation and Competition \\
\textsuperscript{2}Technical University of Munich\\
\textsuperscript{3}inovex GmbH \\
\texttt{\small\{gerhard.hagerer,hannah.danner,archishman.roy,jiaxi.zhao\}@tum.de,grohg@mytum.de}\\
\texttt{\small\{martin.kirchhoff,robert.pesch\}@inovex.de,mainak.ghosh@ip.mpg.de}
}

\date{}

\begin{document}
\maketitle
\begin{abstract}
Recent research in opinion mining proposed word embedding-based topic modeling methods that provide superior coherence compared to traditional topic modeling. In this paper, we demonstrate how these methods can be used to display correlated topic models on social media texts using SocialVisTUM, our proposed interactive visualization toolkit. It displays a graph with topics as nodes and their correlations as edges. Further details are displayed interactively to support the exploration of large text collections, e.g., representative words and sentences of topics, topic and sentiment distributions, hierarchical topic clustering, and customizable, predefined topic labels. The toolkit optimizes automatically on custom data for optimal coherence. We show a working instance of the toolkit on data crawled from English social media discussions about organic food consumption. The visualization confirms findings of a qualitative consumer research study. SocialVisTUM and its training procedures are accessible online\footnote{\url{https://github.com/ghagerer/SocialVisTum}}. 
\end{abstract}

\section{Introduction}

Web sources, such as social networks, internet forums, and customer reviews from online shops, provide large amounts of unstructured text data. Along with the steady development of new platforms and the increasing number of internet users, the interest in methods that automatically extract the expressed opinions along with the corresponding topics and sentiments in text data has increased in recent years. Scholars and organizations from different fields can utilize such methods to identify patterns and generate new insights. Examples are opinion researchers investigating current opinions on political and societal issues, consumer researchers interested in consumers' beliefs about the consumption and production of goods \cite{danner2020combining}, and marketing managers curious about the public perception of their products and services \cite{Berger2020,Murphy2014}. \cite{martinthesis} 

\begin{figure}[t]
\centering
\includegraphics[width=.5\textwidth]{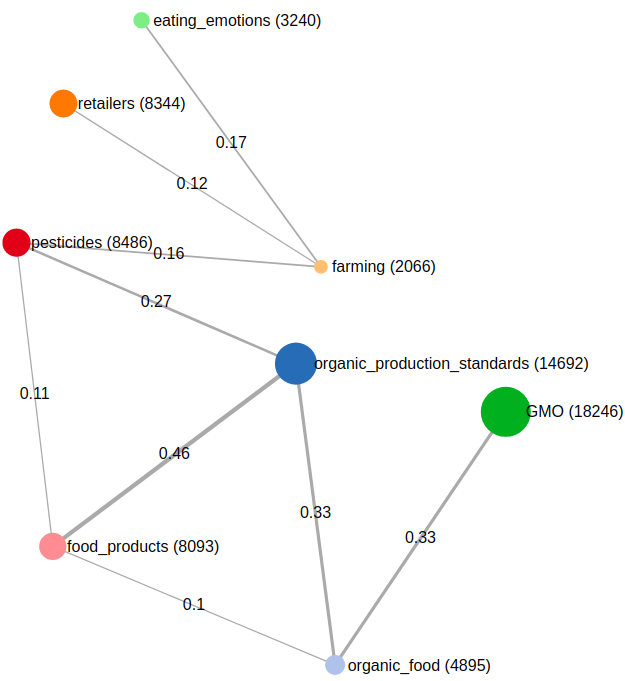}
\caption{SocialVisTUM applied to our use case \textit{organic food} - The topics, their occurrence (in brackets) and respective correlations.}
\label{fig:topic-graph}
\end{figure}

These domain-specific use cases are of interest for research disciplines which taken by itself are not directly related to natural language processing (NLP). Consequentially, there is a constant need to provide state-of-the-art NLP methods such that domain researchers from other fields can take advantage of them. The requirements therefore are simple usage, automatic hyperparameter optimization, minimal effort for manual labeling of text data, and built-in visualizations to give an abstract overview of the discussed topics and their relation with each other.

While these practical requirements are important for domain experts, modern opinion mining approaches target specific machine learning objectives. Recently, there is a trend towards unsupervised neural methods for opinion target detection. Attention-based aspect extraction (ABAE) enables clustering of short review texts with significantly higher coherence as traditional LDA-based topic modeling, and it gives $70\%$ F1 score for classification \cite{he2017unsupervised}. This is improved recently \cite{journals/corr/abs-1909-00415,journals/corr/abs-1808-08858,conf/ijcai/LuoASLYHY19}, which underlines the recent impact and potential of related techniques. 


However, these have not been utilized for visualizations based on correlated topic modeling \cite{blei2006ctm}, where all pairs of topics "are" analyzed to determine if two topics generally tend to occur in the same texts of a given dataset.
Thus, the similarity between topics can be defined. This is successfully used to connect topics (nodes) among each other based on their correlations (edges) leading to more abstract and more meaningful meta topics (graph-clusters) which additionally improves topic coherence. Consequentially, these meta topics, e.g., company-related events or research sub-disciplines \cite{conf/ieeevast/LiuWCZG14,journals/corr/MaiyaR14}, can be successfully identified by graph-based visualization techniques. However, there is a lack of related prototypes on texts discussing consumption related issues in product reviews or social media. To the best of our knowledge, there is also no related integration of sentiment analysis into a system available for potential end users, i.e., domain experts. As according text data from customers is available on a large scale in social media, this can be considered as a shortcoming in the field.





To address all denoted issues, we propose the \textit{SocialVisTUM} toolkit, a new visualization and labeling tool to give users a comprehensible overview of the topics discussed in social media texts. It integrates a neural method for unsupervised sentence and comment clustering based on word vectors and attention. We denote the respective clusters as topics hereafter. In addition, we provide a graph-based visualization showing the topics as labeled nodes and the correlation between them as edges. A force-directed graph layout maintains readability even while many relevant topics and topic relations are displayed. \cite{martinthesis} 

In our interactive graphical user interface, the number of  topics displayed and the correlation threshold required to display a connection between two topics can be dynamically adjusted. Further, contextual topic information is provided, such as the number of respective topic occurrences in the social media texts as node diameter, the correlation between the topic occurrences as edge width, example sentences from the data for each topic, a list of representative words for each topic, and the regarding sentiment distribution of a topic. It is a common practice to represent topics merely by word lists \cite{blei2003latent,chen2014aspect}, which tend to be insufficient to comprehensively express a topic on our given dataset. \cite{martinthesis} 

To avoid manual labeling and to give users an immediate impression of each topic, topic labels are generated automatically based on a custom algorithm utilizing the most common WordNet hypernym in a topic's top words. Furthermore, we find that topic hypernym statistics can serve as a metric for automatic hyperparameter optimization, which in our case gives practical advantages over widely used coherence scoring metrics.

In addition to a more detailed description of our SocialVisTUM toolkit, we show the results of a case study based on social media texts from online commenters debating about organic food consumption. We demonstrate that the correlated topics give a meaningful graph representation of the social media discussions supporting the understanding of the concerns of consumers. In this regard, we also show how the combined illustration of different types of relevant topic and sentiment information and automatic labeling of clusters are a contribution.


\section{Related Work}


Correlated topic models were introduced 2006 \cite{blei2006ctm,10.1145/1143844.1143917} to improve topic coherence and to provide graph visualizations based on topics as nodes and their correlations as edges. This shows potential to improve text mining for the end user as \enquote{powerful means of exploring, characterizing, and summarizing large collections of unstructured text documents} \cite{journals/corr/MaiyaR14}. Meta topics, such as research domains and their inter-disciplinary overlaps, can thus be described clearly, automatically, and empirically \cite{blei_lafferty07}. 


These correlated topic models are applied for more sophisticated visualization approaches. \textit{TopicPanorama} models technology-related topics from various text corpora, including newspaper articles, blogs, and tweets \cite{conf/ieeevast/LiuWCZG14}. Here, the domain expert is given the option to interactively modify the matching result of the labeled topic graph. Another topic visualization called \textit{topic similarity networks} is particularly addressing the visualization of large document sets \cite{journals/corr/MaiyaR14}. While claiming good scalability regarding the number of documents, beneficial methods to achieve automatic topic labeling are successfully quantified. \textit{TopicAtlas} provides a graphical user interface to explore text networks, such as hyperlinked webpages and academic citation networks. For manual mining purposes, topic models are generated and related to one another to facilitate manual navigation and finding of relevant documents \cite{conf/icdm/HeHLSJW16}. These examples show a steady, meaningful, and promising development regarding the visualization of correlated topic modeling, partially also applied to social media texts such as micro-blogs. However, these examples do not include sentiment analysis as means to conduct market research and quantify customer satisfaction in specific and not yet explored market domains. Furthermore, the widely used latent Dirichlet allocation (LDA) technique tends to be incoherent on short texts, such as, product reviews or social media comments, and thus insufficient to detect opinion targets in an unsupervised manner \cite{he2017unsupervised}.




Automatic topic coherence optimization can be seen as desireable for a topic modeling visualization toolkit such as SocialVisTUM, which tries to minimize manual optimization efforts for non-technical users. Therefore, we refer to two widely used coherence definitions \cite{mainakThesis}. Firstly, word co-occurrence-based methods measure how often pairs of representative topic words co-occur in the training data set or in an external reference data set. In that regard, it has been shown that the evaluation methods UMass, UCI and NPMI correlate with human judges 
\cite{stevens2012exploring,newman2010automatic,mimno2011optimizing,bouma2009normalized,ding2018coherence} and are considered to be a default metric for topic coherence.
Secondly, word embedding similarity based coherence scores are recently utilized as these are also based on word co-occurrence statistics \cite{pennington2014glove} and behave similar to NPMI coherence scoring \cite{ding2018coherence}, resulting in high correlation with human perception. These methodologies show the undesirable effect of no distinct local optimum when the hyperparameters of the models are changed, e.g., number of clusters or vocabulary size. On our data, these parameters increase together with the coherence scores, while the subjective performance, i.e., the human perception, actually decreases. We describe this effect and our solution in the case study section.

\section{Clustering Architecture}

The unsupervised neural network model called attention-based aspect extraction (ABAE) \cite{he2017unsupervised} clusters sentences based on GloVe word embeddings \cite{pennington2014glove} and attention \cite{bahdanau2014neural} to focus on the most important words in a sentence. Every sentence $s$ is represented by a vector $z_s$ that is defined as the weighted average of all the word vectors of that sentence. The weights are attentions calculated based on the contribution of the respective words to the meaning of the sentence and the relevance to the topics. These topics are defined by the actual centroids. In their publication, the topics are mapped to aspect classes for unsupervised aspect extraction, which we do not do for our case. \cite{martinthesis}


The topics are initialized as the resulting centroids of k-means clustering on the word embeddings of the corpus dictionary. These are then stacked as topic embedding matrix \textbf{T}. During training, ABAE calculates sentence reconstructions $\mathbf{r}_{s}$ for each sentence. These are linear combinations of the topic embeddings from \textbf{T} and defined as

\begin{equation}
\mathbf{r}_{s}=\mathbf{T}^{\top} \cdot \mathbf{p}_{t},
\end{equation}

where $\mathbf{p}_{t}$ is the weight vector over \textit{K} topic embeddings. Each weight corresponds to the probability that the input sentence belongs to the associated topic.
$\mathbf{p}_{t}$ is obtained by reducing the dimension of $\mathbf{z}_{s}$ to the number of topics \textit{K} and applying softmax such that

\begin{equation} \label{eq:pt}
\mathbf{p}_{t}=softmax\left(\mathbf{W} \cdot \mathbf{z}_{s}+\mathbf{b}\right),
\end{equation}

where $\mathbf{W}$ and $\mathbf{b}$ are trainable and matrix weights and a bias vector respectively. The topic embeddings $T$ are updated during training to minimize the reconstruction error $J(\theta)$ between $\mathbf{r}_{s}$ and $\mathbf{z}_{s}$ based on the contrastive max-margin objective function.
Since words and topics share the same dimensionality, cosine similarity between both can be used to look up the most similar words representing each topic, similar to the way LDA \cite{blei2003latent} represents topics as word distributions. \cite{martinthesis}

\begin{figure}[t]
\centering
\includegraphics[width=.58\textwidth]{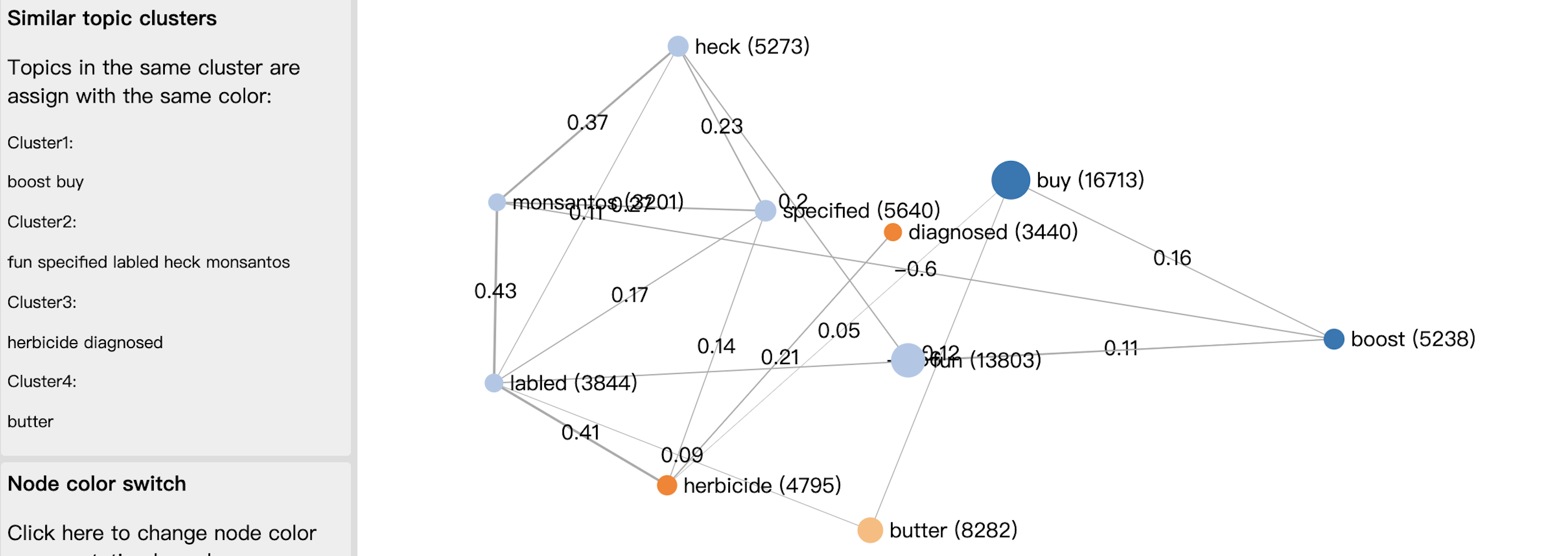}
\caption{Highly correlated topics are colored by the same color respectively.}
\label{fig:hier-clustering}
\end{figure}

\section{The SocialVisTUM Toolkit} \label{sec:toolkit}

\paragraph{Visualization}
Figure \ref{fig:hier-clustering} shows an example of the visualization and labeling tool. Topics are represented as nodes with according labels, and the number of texts assigned to the topics is given in parentheses next to the label. The node size increases based on the number of \textit{topic occurrences}. The edges of the topic connections are labeled by the \textit{topic correlations}. The link thickness increases with a higher positive correlation. A graph layout based on repelling forces between nodes helps to avoid overlaps, which is especially useful when many nodes and links are displayed. A second force keeps the graph centered. Users can also move nodes around to get a more comprehensible overview. \cite{martinthesis}

\paragraph{Topic Nodes and Correlations}

After training the ABAE model, the sentences are assigned to topics based on the maximum topic probability from $\mathbf{p}_{t}$, see formula \ref{eq:pt}. The correlation between two topics $i$ and $j$ is calculated based on the probabilities $\left( \mathbf{p}_{t} \right)_i$ and $\left( \mathbf{p}_{t} \right)_j$ of each given sentence $t$. This yields a value in the range of $[-1, 1]$ for every pair of topics specifying the strength of the corresponding relatedness. \cite{martinthesis} 



\paragraph{Hiding Insignificant Topics}

An occurrence threshold slider defines the percentage of sentences that must be about a topic to display the associated node. Another slider can be used to set the correlation threshold to define the required positive or/and negative correlation to display the associated connections. These sliders are especially helpful to maintain a clear visualization by limiting the number of shown topics and connections when there are many of them available.

\begin{figure}[t]
\centering
\includegraphics[width=.36\textwidth]{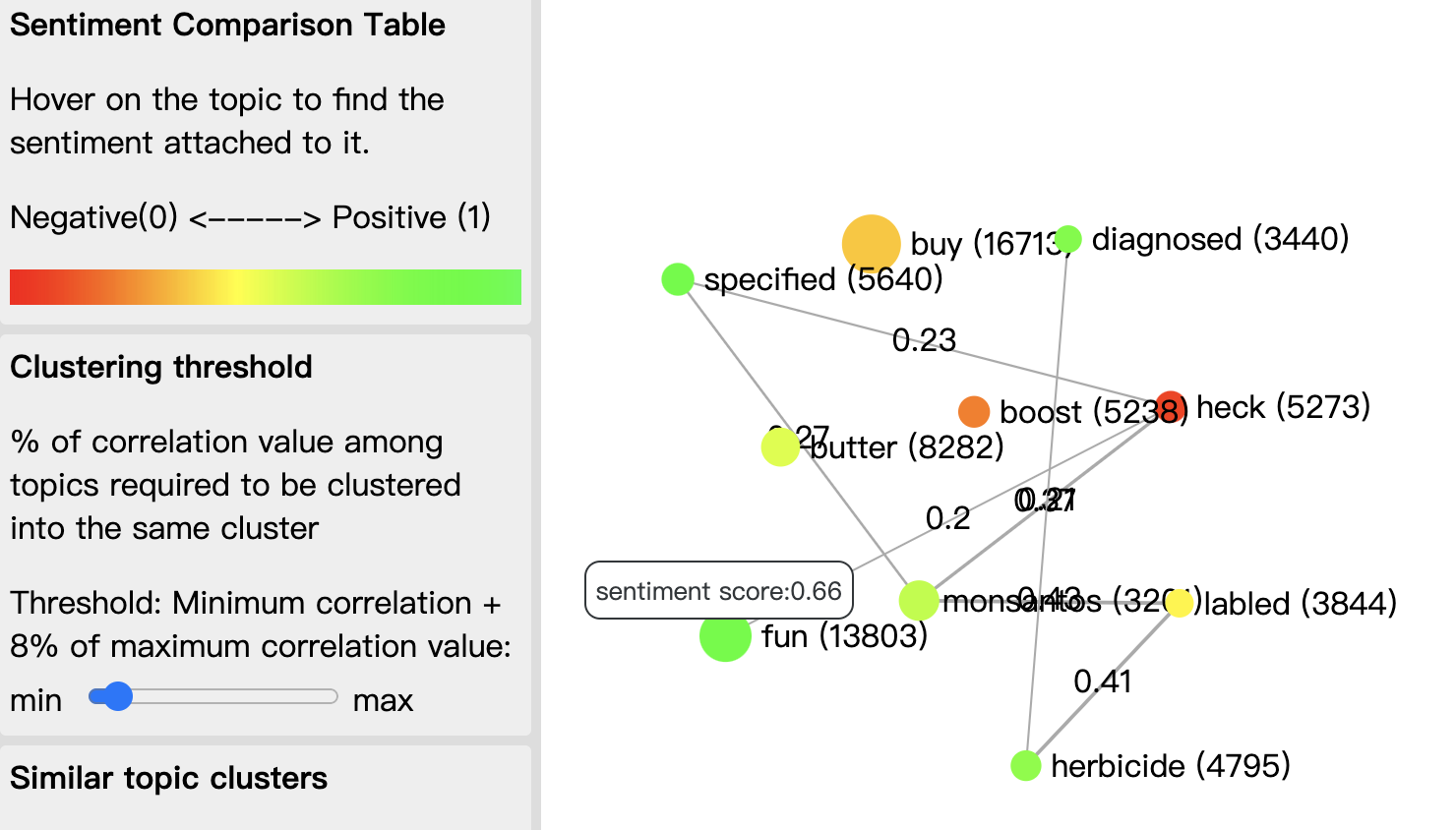}
\caption{The sentiment for each topic is shown as green (positive) and red (negative).}
\label{fig:sent-analysis}
\end{figure}

\paragraph{Topic Inspection} \label{par:topic-inspection}

Users can double-click a node to receive additional information about a topic, i.e., representative words and sentences, as shown in Figure \ref{fig:15C_manual-labels_pesticides_complete} on the left and right respectively. As representative words, the top 10 words are shown sorted by the distance of their embeddings to the selected topic centroid in ascending order. The representative sentences are the ones with the highest probability from $p_t$ for the given topic. During topic inspection mode, only nodes that are connected to the clicked node and the associated links are displayed. A double click on the same node brings back the whole graph again.

\begin{table*}[t]
\centering
\begin{tabularx}{0.9\textwidth}{
    >{\hsize=1\hsize}r  
    >{\hsize=1.4\hsize}l 
    >{\hsize=.6\hsize}r}
\hline
\textbf{Topic Label} & \textbf{Representative words} & \textbf{\# Hypernyms}\\
\hline
animal (102) & insect, ant, habitat, rodent, herbivore & 218\\
compound (91) & amino, enzyme, metabolism, potassium, molecule  & 158\\
chemical (74) & fungicide, insecticide, weedkiller, preservative, bpa & 131\\
systematically (0) & systematically, adequately, cleaned, properly, milked & 0 \\
\hline
\end{tabularx}
\caption{Example topics, automatically assigned topic labels, and representative words. The value next to the topic label denotes how often the label occurs as a shared hypernym. The number of hypernyms on the right tells in how many word comparisons a shared hypernym is identified. Taken from \cite{martinthesis}.}
\label{tab:vocab_10000_topics_50}
\vspace{1.5em}
\end{table*}

\paragraph{Colorization of Topic Nodes}

In an updated version of SocialVisTUM, we introduce two meaningful colorings of the topic nodes for correlated topic clustering and sentiment analysis.

Firstly, we perform a hierarchical clustering algorithm such that those topics which are strongly correlated with each other are colorized in one and the same color respectively. A dynamic slider GUI element helps to adjust the correlation threshold accordingly. One example outcome is shown in \ref{fig:hier-clustering}.

Secondly, we perform sentiment analysis using the Valance Aware Dictionary and sEntiment Reasoner or VADER method \cite{gilbert2014vader}. It is a lexicon and rule-based sentiment analysis tool that is specifically attuned to sentiments expressed in social media settings. It gives positive, negative, and neutral scores to represent the proportion of text in that sentence that falls in these categories. For each sentence, we use the compound score, i.e., the sum of all lexicon ratings normalized between -1 (most negative) and +1 (most positive). We then calculate the average sentiment score for each topic based on all respective topic sentences. In Figure \ref{fig:sent-analysis}, positive sentiment is shown as green topic nodes, and negative as red. \cite{archieJiaxiLabReport}

\paragraph{Automatic Topic Labels} \label{par:autom-topic-labels}

\begin{figure}[t]
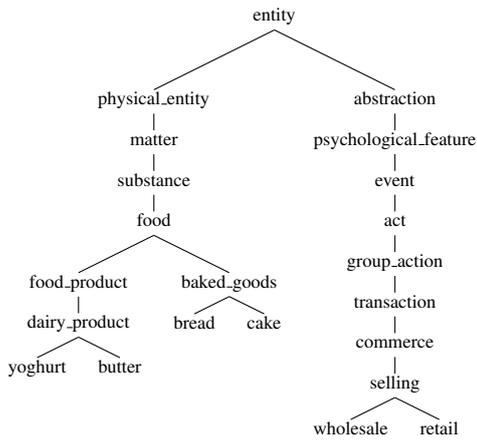

\centering
\scriptsize
\Tree[.entity 
[.{physical\_entity} [.matter [.substance [.food 
[.{food\_product} [.{dairy\_product}
yoghurt 
butter ]
] 
[.{baked\_goods} bread cake ] ]]]
]
[.abstraction [.{psychological\_feature} [.event [.act [.{group\_action} [.transaction [.commerce [.selling 
wholesale 
retail ] ]]]]]]
]
]
\caption{Hypernyms for \textit{yoghurt}, \textit{butter}, \textit{bread}, \textit{cake}, \textit{wholesale}, and \textit{retail}. Taken from. \cite{martinthesis}}
\label{fig:hypernym_path_example}
\vspace{1em}
\end{figure}

We introduce an approach to label topic nodes automatically. It is based on \textit{shared hypernyms}, i.e., the lowest common denominator for words, which we identify using the representative topic words denoted in the previous paragraph and the lexical database WordNet \cite{miller1995wordnet}. First, we retrieve the hypernym hierarchy for every representative topic word, as shown in Figure \ref{fig:hypernym_path_example}, and compare every word with every other word in the word list. Next, at each comparison, we save the hypernym with the lowest distance to the compared words in the hypernym hierarchy. We denote these as \textit{shared hypernyms}. We only consider hypernyms if their distance to the word is smaller than half of the distance to the root hypernym to avoid unspecific labels like \textit{entity} and \textit{abstraction}. Eventually, we employ the hypernym that occurs most often as topic labels. If no hypernym can be identified, we use the most representative word instead. In the example shown in figure \ref{fig:hypernym_path_example}, we identify \textit{dairy\_product} as the lowest shared hypernym of \textit{yoghurt} and \textit{butter}, and \textit{food} as lowest shared hypernym of \textit{yoghurt} and \textit{bread}. \cite{martinthesis}

The quality of a shared hypernym chosen as topic label can be approximated by inspecting the number of its hypernym occurrences -- see table \ref{tab:vocab_10000_topics_50}. Topic labels occurring frequently as shared hypernym are usually suitable (e.g., animal (102) and compound (91)) in contrast to topic labels that rarely occur (e.g, group\_action (9) or smuckers (0)). Thus, we conclude that the number of hypernym occurrences of each topic is suitable to estimate the topic coherence for hyperparameter optimization -- see section \ref{sec:hyperparam} later on. \cite{martinthesis}

\paragraph{Changing Labels}

To change the label of a topic, the user can click on the associated label of a node. This opens a prompt allowing the user to change the topic label. The user can download a JSON file with the updated labels by clicking on the \textit{Create file} button on the sidebar. \cite{martinthesis}

\section{Case Study}

We demonstrate SocialVisTUM's potential for social media data exploration on a new data set. 

\subsection{Data Set}

We crawled online user comments on organic food from multiple forums (e.g., Reddit, Quora, Disqus) and the comment sections of news websites (The Washington Post, The New York Times, Chicago Tribune, HuffPost, and many more). The goal is to discover the discussed topics and opinions in social media regarding the organic food consumption. 

Relevant articles from the platforms are found by the search terms "organic food", "organic agriculture", and "organic farming". We further filtered for domain relevance by applying naive Bayes classification on bag of words trained on 1000 random and accordingly labeled texts (84.70\% accuracy with 10-fold cross validation). From the left texts, we retain comments containing either of the words \textit{food} and \textit{organic}. The left data set consists of 515.347 sentences totaling 83.938 posts, which are used to train the ABAE topic model. We use the 300-dimensional pre-trained GloVe embeddings and fine-tuned them on the data. \cite{martinthesis} 


\subsection{Hyperparameter Estimation} \label{sec:hyperparam}

Some hyperparameters of the utilized ABAE topic model are the number of topic clusters and the vocabulary size. To optimize these automatically, we define a new metric, the \textit{average number of shared hypernyms} (ANH). We first derive the frequency of all shared hypernyms for each topic, which is already done for automatic topic labeling in section \ref{par:autom-topic-labels}. The ANH is the sum of hypernym frequencies over all topics divided by the number of topics. \cite{martinthesis} 

In our case study, we identified the following advantages of ANH over the widely used coherence score (CS). First, an increasing number of topics does not always increase the ANH, as a high number of topics leads to many incoherent topics with fewer shared hypernyms, i.e., a lower ANH. Second, a medium-sized vocabulary (\texttildelow 10,000 words) produces the most coherent topics according to ANH and manual inspection. Table \ref{tab:result_comparison} shows an excerpt of the results for varying parameters.

\subsection{Interpretation}

We applied SocialVisTUM to our case of organic food yielding the topics displayed in figure \ref{fig:topic-graph}. A consumer researcher in the domain of organic food manually refined the automatic labeling based on the most similar words of each topic. The topics reflect previous findings of a qualitative content analysis on a small sub-sample of our data set \cite{Danner2020}. The correlated topics allow market researchers to investigate the context in which topics are discussed.

\begin{figure}[t]
\centering
\includegraphics[width=.49\textwidth]{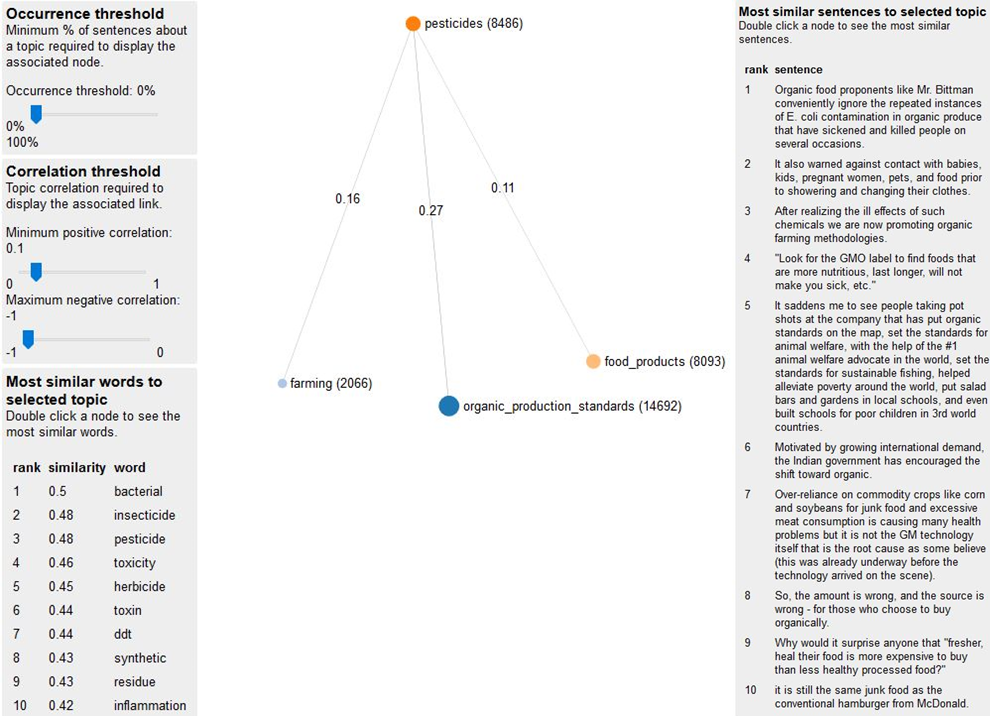}
\caption{SocialVisTUM applied to our \textit{organic food} use case. Topic inspection of the \textit{pesticides} topic.
}
\label{fig:15C_manual-labels_pesticides_complete}
\vspace{1em}
\end{figure}

Figure \ref{fig:15C_manual-labels_pesticides_complete} takes a closer look at the example topic \textit{pesticides}, which is concerned with different pesticides and their toxicity. The topic \textit{pesticides} is correlated with the topic \textit{organic\_production\_standards}, which references different organic or related production methods, such as bio-dynamic, hydroponic, or bio-intensive agriculture. This  correlation suggests that, for the commenting users in our data set, the non-use of chemical-synthetic pesticides is an important characteristic of organic compared to non-organic production. Further topics correlated with \textit{pesticides} propose that the commenting users are concerned about the use of \textit{pesticides} in \textit{farming} and that they discuss the issue of \textit{pesticides}, possibly the residues thereof, in the context of different \textit{food\_products}.

\section{Conclusion}

In this paper, a case of the proposed SocialVisTUM demonstrates the visualization of coherent topics on a given corpus of social media texts about organic food. The graph-based visualization with topics as nodes and topic correlations as edges reflects the topics and patterns found in a related qualitative content analysis \cite{Danner2020}. The presentation of additional topic information, such as word lists, representative sentences, topic importance, and meaningful predefined labels, provide a basis for the understanding and interpretation of a topic for domain experts. The integrated hyperparameter optimization automatically yields interpretable topics and helps tailoring the model to the given data set. For future work, we plan to evaluate the correlated topics on other corpora and in other use cases. In addition to Pearson correlation, other correlations could improve the approach. We plan to integrate multi-lingual word features, such as BERT \cite{devlin2018bert}, for cross-cultural comparisons.


\section*{Acknowledgments}

We thank inovex GmbH for supporting the research of this paper by providing computational resources, Robert Pesch and Martin Kirchhoff for their important contributions. Paragraphs adopted from student works are followed by the corresponding citation after the punctuation.

\newpage

\bibliographystyle{acl_natbib}
\bibliography{anthology,ranlp2021}

\appendix

\newpage

\section*{Appendix A. Comparison of Coherence Metrics}

\begin{table}[h]
\centering 
\begin{tabularx}{0.82\linewidth}{c|r|r|r}
\hline
\textbf{\# Topics} & \textbf{Voc. Size} & \textbf{CS} & \textbf{ANH} \\ 
\hline		
\multirow{3}{*}{5} & 1,000 & -1104 & 28.6 \\
& \textbf{10,000} & -765 & \textbf{68.0} \\
& 18,000 & -403 & 5.2 \\
\hline
\multirow{3}{*}{15} & 1,000 & -366 & 33.3 \\
& \textbf{10,000} & -270 & \textbf{40.0} \\
& 18,000 & -197 & 33.8 \\
\hline
\multirow{3}{*}{50} & 1,000 & -110 & 30.4 \\
& \textbf{10,000} & -70 & \textbf{51.8} \\
& 18,000 & -54 & 49.7 \\
\hline
\end{tabularx}
\caption{Comparing two coherence metrics: coherence score (CS) and average number of shared hypernyms (ANH). The advantage of ANH is that it has its global optimum always in the middle as opposed to CS. This property is beneficial for hyperparameter optimization.}
\label{tab:result_comparison}
\end{table}

\end{document}